\documentclass[12pt,journal,compsoc,onecolumn]{IEEEtran}
\ifCLASSOPTIONcompsoc
  \usepackage[nocompress]{cite}
\else
  \usepackage{cite}
\fi

\usepackage{cite}

\ifCLASSINFOpdf
\else
\fi

\usepackage{graphicx}
\usepackage{amsfonts}
\usepackage{amsmath}
\usepackage{makecell}
\usepackage{xcolor}

\usepackage[numbers]{natbib}

\usepackage{setspace}
\usepackage{longtable}
\usepackage{afterpage}
\usepackage{booktabs}
\usepackage{multirow}
\begin{document}
\title{Specialized Re-Ranking: A Novel Retrieval-Verification Framework for Cloth Changing Person Re-Identification}

\author{Renjie~Zhang,
Yu~Fang,
Huaxin~Song,
Fangbin~Wan,
Yanwei~Fu,
Hirokazu~Kato,
and~Yang~Wu
\IEEEcompsocitemizethanks{\IEEEcompsocthanksitem R. Zhang, Y. Fang, and H. Kato are with the Division of Information Science, Nara Institute of Science and Technology, Nara, Japan (Email:zhang.renjie.zq0@is.naist.jp; fang.yu.fq0@is.naist.jp; kato@is.naist.jp).}
\IEEEcompsocitemizethanks{\IEEEcompsocthanksitem H. Song is with the Department of Mathematical Informatics, Tokyo University, Japan(e-mail: song-h@g.ecc.u-tokyo.ac.jp).}

\IEEEcompsocitemizethanks{\IEEEcompsocthanksitem F. Wan and Y. Fu are with the School of Data Science, Fudan University, Shanghai, China((e-mail: fbwan18@fudan.edu.cn; yanweifu@fudan.edu.cn).}
\IEEEcompsocitemizethanks{\IEEEcompsocthanksitem Y. Wu is with the ARC Lab, Tencent PCG, Shenzhen, China (email: dylanywu@tencent.com).}
}

\IEEEtitleabstractindextext{%
\begin{abstract}

\linespread{1.25}
Cloth changing person re-identification(Re-ID) can work under more complicated scenarios with higher security than normal Re-ID and biometric techniques and is therefore extremely valuable in applications.  Meanwhile, higher flexibility in appearance always leads to more similar-looking confusing images, which is the weakness of the widely used retrieval methods. In this work, we shed light on how to handle these similar images. Specifically, we propose a novel retrieval-verification framework. Given an image, the retrieval module can search for similar images quickly. Our proposed verification network will then compare the input image and the candidate images by contrasting those local details and give a similarity score. An innovative ranking strategy is also introduced to take a good balance between retrieval and verification results. Comprehensive experiments are conducted to show the effectiveness of our framework and its capability in improving the state-of-the-art methods remarkably on both synthetic and realistic datasets.
\end{abstract}

}

\maketitle
\IEEEdisplaynontitleabstractindextext
\IEEEpeerreviewmaketitle

\IEEEraisesectionheading{\section{Introduction}\label{sec:introduction}}

\IEEEPARstart{P}{erson} re-identification (Re-ID) aims to find out images of the same person across multiple images taken in different scenes. Essentially, Re-ID is a kind of unobtrusive identification. It neither collects detailed biological traits such as fingerprints and facial appearance nor requires the cooperation of users, which makes differences from traditional identification methods. It is also an extremely valuable computer vision task for its use in public safety and security area such as person localization, multi-object tracking, and video surveillance. Recently proposed benchmarks \cite{MTMC01,mar01} and works \cite{PLR01,Sphere01,PCB01} have greatly promoted the research in this community. However, these methods are limited by a strong assumption that the same people do not change their clothes. We categorize such models as the cloth-consistent Re-ID models. 

In contrast, cloth changing Re-ID models aim to recognize the person of interest under drastic appearance changes, which is another important component of a Re-ID system. Compared with the cloth-consistent Re-ID models, cloth changing Re-ID models can work in more complicated scenarios for a longer time span while with a lower risk of data abuse since identification by human eyes is also harder. These advantages make cloth changing Re-ID models more effective and practical in various applications. On the other hand, higher appearance flexibility also inquires models to make the correct selection from a large number of similar candidate images. Since the person of interest is no longer limited by his/her color appearance, more candidate images that may share similar gestures, viewpoints, or body shapes are collected in the dataset. Distinguishing these similar images is always a tough task even by humans, not to mention a model that is not specially designed for this. Nevertheless, we find that if we can learn how these similar images differ, it will help the decision-making among vast similar images. Unfortunately, many researchers have underestimated the value of this task as well as the importance of details in this problem. Inspired by this, we try to solve this potential and challenging problem by better analyzing those local details.

The biggest challenge in cloth changing Re-ID is that there are no features as intuitive and decisive as color appearance. The system should pay more attention to other features like body shapes, faces, and gestures than before. Relying too much on one specific type of features will lead to a biased result. Generally speaking, such auxiliary features are not that representative and can only handle some easy works\cite{LTCC01,cc02}. Following this idea, many cloth changing Re-ID works\cite{LTCC01,Part01,VC01} devotes their efforts to incorporating various auxiliary features, especially body shape features. Unfortunately, these works fail to use these features in a correct manner. Almost all these works take auxiliary features as a kind of general features(i.e. features given by retrieval models). Though such design is common in cloth-consistent Re-ID, it helps little when it comes to cloth changing Re-ID(see \textcolor{blue}{Table \ref{VC1}} and \textcolor{blue}{Table \ref{PRCC1}}). In cloth-consistent Re-ID, candidate images usually share similar color appearances but with remarkable differences in other aspects such as body shape and accessories. In this case, auxiliary features can help eliminate some unwanted images and bring a better result. In contrast, candidate images in cloth changing Re-ID datasets are always hard to discern even by human eyes. They usually share similar global features with just some inconspicuous distinctions in local parts. Auxiliary features are typically not representative enough for discerning these hard samples. 

To solve this problem, the key point is to re-activate auxiliary features. We notice that the malfunction of auxiliary features is caused by the difference in data distribution. For example, supposing auxiliary features are trained to tell whether a person is strong or not, but when analyzing the similar candidate images, we are comparing two strong people with tiny differences. Such difference in data distribution causes the performance reduction of auxiliary features. To re-activate auxiliary features, we should make them representative enough for distinguishing those hard samples. One nature idea is to learn auxiliary features that can discern similar images. To reach this target, one should train auxiliary features to be high-granularity. On the other hand, learning such all-mighty features with limited samples is extremely difficult. This restrictions forces us to seek other possibilities. Alternatively, we propose verifying similar images through auxiliary features, which mainly differs from the former method in two aspects:  1). The target is changed to verify whether two images are from the same person. This is a more directive way of fulfilling the responsibility of auxiliary features, i.e. eliminating unwanted images through specific features. The verification score is adopted for inference. Compared with feature distance, the verification score given by the network can better balance the contribution of different auxiliary features and is thus more stable and reliable. 2). Auxiliary features are jointly decided by two input images after adequate information exchange. Such a setting enables the model to flexibly adjust its expression according to different combinations of the input images. In other words, features of the same image are dynamically changing according to different images to be compared with. We categorize this kind of features as specialized features. With the help of specialized features, the system can actively and precisely detect the important clues in the input image pair and produce more representative features. 

In this work, we introduce a novel retrieval-verification scheme for solving this problem. Instead of using general features extracted by Re-ID models to directly give a rank, we utilize these features for finding a shot list of reliable candidate images in the retrieval part of the scheme. These candidate images typically have inconspicuous while decisive differences in various points. Then, a specially designed verification network is proposed to learn the specialized features of these candidate images. Different from the retrieval part, the verification network takes a pair of candidate image and query image as its inputs and outputs a similarity score of the image pair. Such a setting forces the verification network to automatically detect, judge, and balance the similarity of detailed local image parts. The combination of image retrieval and verification makes proper use of both general features and specialized features. Intuitively, general features given by the retrieval model provide a basic sketch to the person of interest, while distinctive details in the image are in fact the ones decisive. In contrast, specialized features are induced jointly by two input images and can better represent the difference between the images. Therefore, we utilize general features in the retrieval part as a high-speed and high-quality similar image filtering. To further distinguish these similar images, we apply an image verification network to explore the specialized features of those inconspicuous yet decisive differences in local parts.

To better contrast the candidate images with the query images, hint locating becomes the key to a successful verification network. In this work, we incorporate human part-based segmentation to parse the image. By parsing the image into human body parts, part-to-part matching can be accomplished automatically. This greatly help narrow the area for hint searching and improve the efficiency of comparison. Another biggest advantage of human part-based segmentation is its pixel-level accuracy. This allows us to analyze the input image pairs from the viewpoint of part shape, which is overlooked by many cloth changing Re-ID works\cite{cc01,PRCC01,LAST01} but is actually very important for hard sample identification. Furthermore, high-order features including the length ratio of limbs and body proportions, are also extremely crucial for image verification. Fusing the above ideas, we propose a novel verification network named Local Clues oriented Verification Network(LCVN) for similar image verification. The structure of our LCVN is shown in \textcolor{blue}{Fig. \ref{Framework}}. 

We also discuss another key point for establishing a successful retrieval-verification framework, which is to know when to stop. Actually, building an all-mighty image verification network is almost impractical, as it works on an exponentially boosted input space. Without enough data and computing power, it is impossible to compare all images for similarity calculation. Including more "dissimilar" images usually leads to more meaningless comparisons and drastic performance drops. To best exert the effectiveness of our verification network, we propose a new ranking strategy for our retrieval-verification framework, which takes a good balance between comparing more images and reducing the risk of performance drops. 

In summary, our contributions are listed as follows:

(1). We propose a novel two-step retrieval-verification strategy to use metric learning results for candidate image filtering and another professional verification network for similar image judgment. To the best of our knowledge, this is the first work that learning specialized features for solving the similar images problem in cloth-changing Re-ID.

(2). A specially designed image verification network named Local Clues oriented Verification Network(LCVN) is proposed to learn the specialized features of similar images in cloth changing Re-ID. It takes both visual clues and inter-parts relationships for giving comprehensive comparisons to similar image pairs. Furthermore, we also introduce a new ranking strategy that can best exert the potential of the verification module.

(3). Comprehensive experiments are given to verify the effectiveness of our model on both the VC-Clothes\cite{VC01} dataset and the PRCC\cite{PRCC01} dataset. Our method improves the state-of-the-art methods by a large margin on both datasets, showing the effectiveness of our proposed framework.

\section{Related Work}
\textbf{Person Re-Identification} Typically, person re-identification is the task of finding out images of the same person across multiple images taken in different scenes. It was the first time when Gheissari et al. \cite{ReIDhis} published their work, Re-ID got separated from multi-camera tracking and become an independent computer vision task. Later on, many works \cite{HOReID01,LOMO01,HOG01} followed this task and established the framework of the Re-ID systems. Recent Re-ID systems mainly consist of three components: feature representation, energy function, and ranking strategy. Almost all recent works\cite{HOReID01,ABD01,Rank01} tried to tackle the challenges in Person Re-ID from three components that are mentioned above. Lin et al. \cite{Lin01} proposed a multi-camera consistent matching constraint to obtain a globally optimal representation in a deep learning framework. In \cite{Sphere01} and \cite{AM01}, the variants of the cross-entropy loss function, such as the sphere loss and AM softmax are investigated. Zhong et al.\cite{Rank01} optimized the ranking order by automatic gallery-to-gallery similarity mining. Through continually improving the model from these three components, satisfying results have been achieved in recent years. However, these works heavily rely on the color appearance for learning universal features, which is typically unavailable under the cloth changing Re-ID setting. Besides, performance drops are observed in \cite{cc02} when applying traditional Re-ID models directly to the cloth changing Re-ID challenge, showing the necessity of a professional model for the cloth changing Re-ID task.

\textbf{Cloth Changing Re-ID approaches} There are only a few works on the cloth changing Re-ID problem\cite{cc02,Part01,VC01,cc01,PRCC01,LAST01,Overall01}. Most of them \cite{LTCC01,VC01,PRCC01,Cele01,Cele02} are proposed with dataset. The common target of these methods is to learn cloth-unrelated features from body shape \cite{cc02,cc01,PRCC01} or faces \cite{VC01}. Yang et al. \cite{PRCC01} introduced a spatial polar transformation on contour sketch to learn shape features. In \cite{LTCC01}, key points of the human body are detected to eliminate the impact of color appearance. In \cite{VC01}, face patches are segmented and analyzed separately. Hong et al.\cite{Part01} shares a similar idea of learning part shape feature for identification. These works take those cloth-unrelated features as a part of general features while failing in checking whether those auxiliary features are representative enough for distinguishing the candidate images in cloth changing Re-ID. In this work, we alternatively adopt cloth-unrelated features for only similar images and train them to be specialized features. Our proposed retrieval-verification framework intentionally incorporates an image verification model to learn the specialized features of two similar images and to estimate their similarity score as an index for ranking. By automatically detecting and judging the significance of those local differences, our proposed verification network can learn more representative auxiliary features for comparing the candidate images than other methods.

\textbf{Verification Models in Re-ID} Verification model is a kind of model that determines whether the input satisfies specific requirements. In some cases, verification model will output a score representing the confidence of the decision. In general, verification models in Re-ID can be classified into two categories. In the early stage of the development of Re-ID research, verification models are mainly adapted from those hand-crafted features based methods\cite{LOMO01,HOG01}. Verification of this style \cite{Veri01} mainly follows the flow of feature extraction, photometric and geometric transforms, and similarity estimation. In this case, the final similarity score produced by the network is used for final ranking. The query image has to be compared with every gallery image before the final ranking. On the other hand, recent metric learning based methods \cite{Veri03} prefer to incorporate a verification module as an additional learning target to train more discriminative features. For example, Zheng et al. \cite{zheng01} included verification loss as well as identification loss to learn a discriminative CNN embedding. In this work, we also include an image verification network. Our proposed framework is mainly different from these works by: 1).  Our image verification network is designed only for similar image pairs. 2). Our image verification is designed to learn and analyze specialized features, while other verification networks only work on verifying general features.

\begin{figure*}[!t]
\centering
\includegraphics[width=\textwidth,height=4.5in]{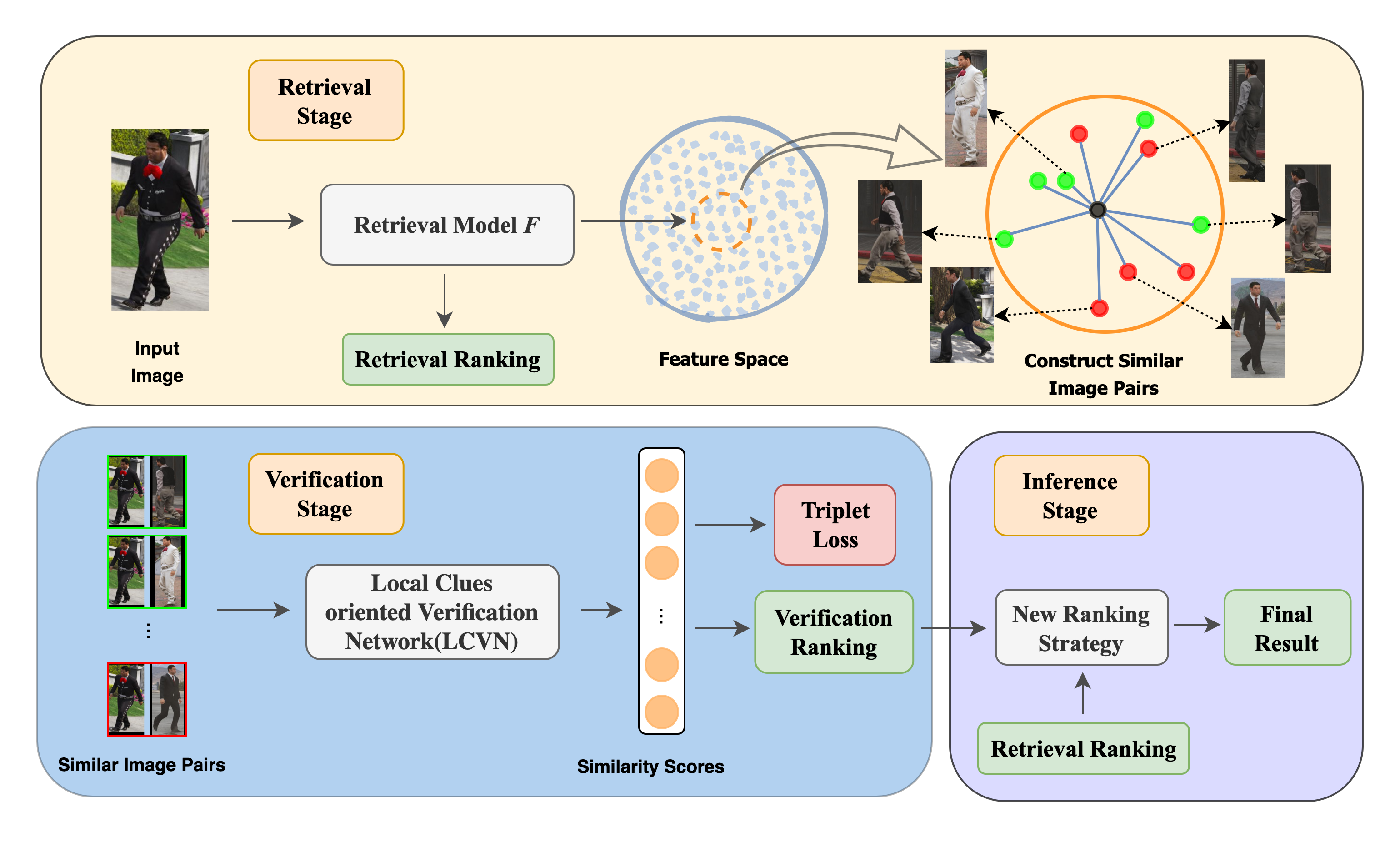}
\caption{\doublespacing An overview of our proposed retrieval-verification framework. In the retrieval stage, we train the retrieval model $F$ following the standard flow. Then we project images into the feature space and construct new similar image datasets following Section 3.1. In the verification stage, we train our verification network LCVN on these similar image pairs to learn specialized features as well as their similarity scores. During inference, both the retrieval ranking and the verification ranking are balanced for the result. Green dots and surrounding boxes denote positive samples. Red dots and surrounding boxes denote negative samples.}
\label{Two-Step}
\end{figure*}

\section{Methodology}

This section will present our framework for the cloth changing Re-ID problem. As mentioned, with large appearance variation, there are more pedestrians with similar appearance, bringing interference to image retrieval. To discriminate these similar images without the loss of efficiency, we propose a novel retrieval-verification framework, which consists of two modules: one retrieval module for finding out similar image pairs and one verification module for comparing them. An overview of the proposed framework is shown in \textcolor{blue}{Fig. \ref{Two-Step}}.

\subsection{Candidate Image Retrieval Module}
A cloth changing Re-ID system usually works on five datasets $\mathbf{T},\mathbf{VQ},\mathbf{VG},\mathbf{Q},\mathbf{G}$, representing training set, validation query set, validation gallery set, query set and gallery set. Let $\mathbf{X}$ be one of the five sets, i.e. $\mathbf{X} \in \{\mathbf{T}, \mathbf{VQ}, \mathbf{VG}, \mathbf{Q},\mathbf{G}\}$. $\mathbf{X}$ contains $N_{X}$ images: $\{X_{i}\}_{i = 1}^{N_{X}}$. Each image is provided with corresponding labels \textit{$X_{i}^y,X_{i}^c \in \mathbb{N}$}, which are its identity label and cloth index label, respectively.

The main purpose of candidate image retrieval module is to find a model $F$ such that it can give a list of images $\mathbf{G_{i}^{F}} = \{G_{i_{1}^{F}}, G_{i_{2}^{F}}, \cdots, G_{i_{P}^{F}}\}$ that are visually similar to image $Q_{i}$ for each $i \in \{1,2,\cdots,N_{Q}\}$. Here $P$ represents the number of candidate images we keep for each query image. We use this model $F$ to prepare the dataset for the second stage of training. For a fair comparison, we evaluate the performance on the validation sets $\mathbf{VQ}$ and $\mathbf{VG}$ when choosing retrieval model $F$. Once the training of the retrieval model $F$ finishes, we build our new validation set $\mathbf{Valid^{F}}$ as follows:

\begin{equation}
\mathbf{Valid^{F}} = \{(VQ_{i}, VG_{i_{k}^{F}},Valid_{i_{k}^{F}}^{y}), i = 1,2,\cdots,N_{VQ}, k = 1, 2, \cdots, P\}
\end{equation}
where $VG_{i_{k}^{F}}$ is the image with the $k$-th biggest similarity with $VQ_{i}$ in $VG$. $Valid_{i^{F}_{k}}^{y}$ is a binary label denoting whether $VQ_{i}$ and $VG_{i_{k}^{F}}$ is the same person. Before establishing new datasets, we exclude all image pairs which share the same identity and same cloth index in advance.

New test set $\mathbf{Test^{F}}$ is established similarly by replacing $\mathbf{VQ}$ and $\mathbf{VG}$ with $\mathbf{Q}$ and $\mathbf{G}$. While for the new train set $\mathbf{Train^{F}}$, in order to take a balance among positive samples and negative samples, we take $P$ most similar same-identity images and $P$ most similar different-identity images separately for each $T_{i}$.

\subsection{Image Verification Module}

\begin{figure*}[!t]
\centering
\includegraphics[width=.9\linewidth]{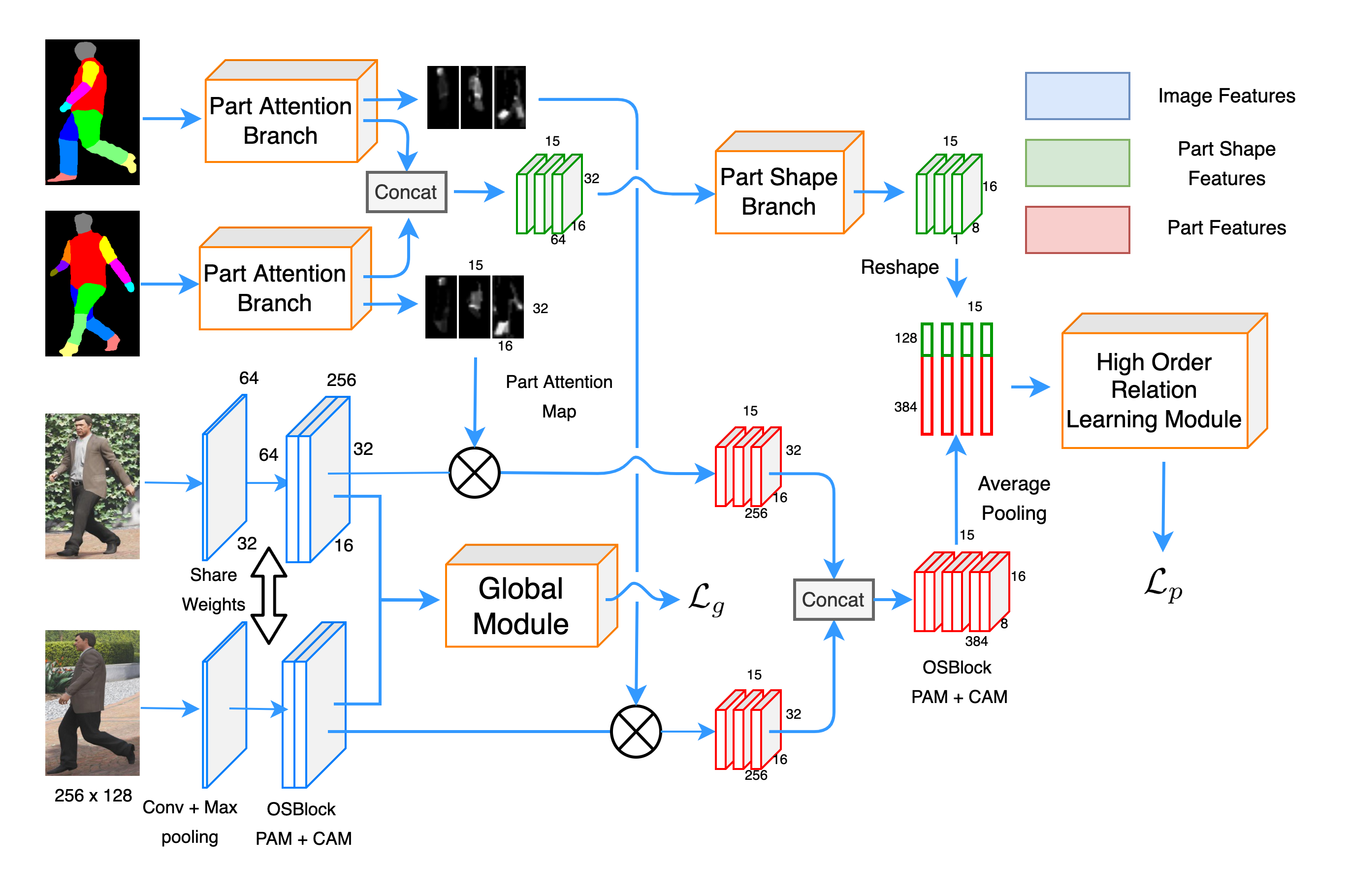}
\caption{\doublespacing The framework of our proposed image verification network. It consists of a global feature learning module $\mathcal{G}$, a part feature learning module $\mathcal{P}$(part attention branch and part shape branch), and a high order relation learning module $\mathcal{H}$. Module $\mathcal{G}$ learns the similarity of two input RGB images directly. Working as the second learning target, it helps us learn discriminative image features for further part feature learning. In $\mathcal{P}$, part attention map and part shape features of each body part are induced from a human parsing map. Fusing both part features and part shape features of two input images, intra-part relationships are modeled from two viewpoints. At last, the high order relation learning module $\mathcal{G}$ is incorporated for revealing those inter-part clues. When inferring, we only take the similarity scores given by $\mathcal{G}$. }
\label{Framework}
\end{figure*}

The candidate image retrieval module only provides a coarse search for some potential candidates. Without representative auxiliary features, current Re-ID models are prone to fail when handling similar images. This limitation is much more obvious when there is no dominant ID-related features as the color appearance features in cloth-consistent Re-ID. To learn representative auxiliary features for similar images, inter-image comparisons are indispensable. Though there are several works\cite{HOReID01,ABD01,zheng01,OS01} that try to reveal the inter-image relationship using verification models, they are usually simple in structure. Wang et al.\cite{HOReID01} shares a similar structure with our verification network, but they still aim at learning general features. Their verification part is very shallow and is designed just for comparing heterogeneous general features. Actually, the key point of a successful verification network for hard samples is specialized features. Specialized features are decided jointly by two images. Information exchange between two images encourages the network to actively and adaptively extract those informative clues and produce more representative features. To learn specialized features, we design our verification network by fusing the features of two input images at an early stage for adequate information exchange.

\begin{figure*}[!t]
\centering
\includegraphics[width=.9\linewidth]{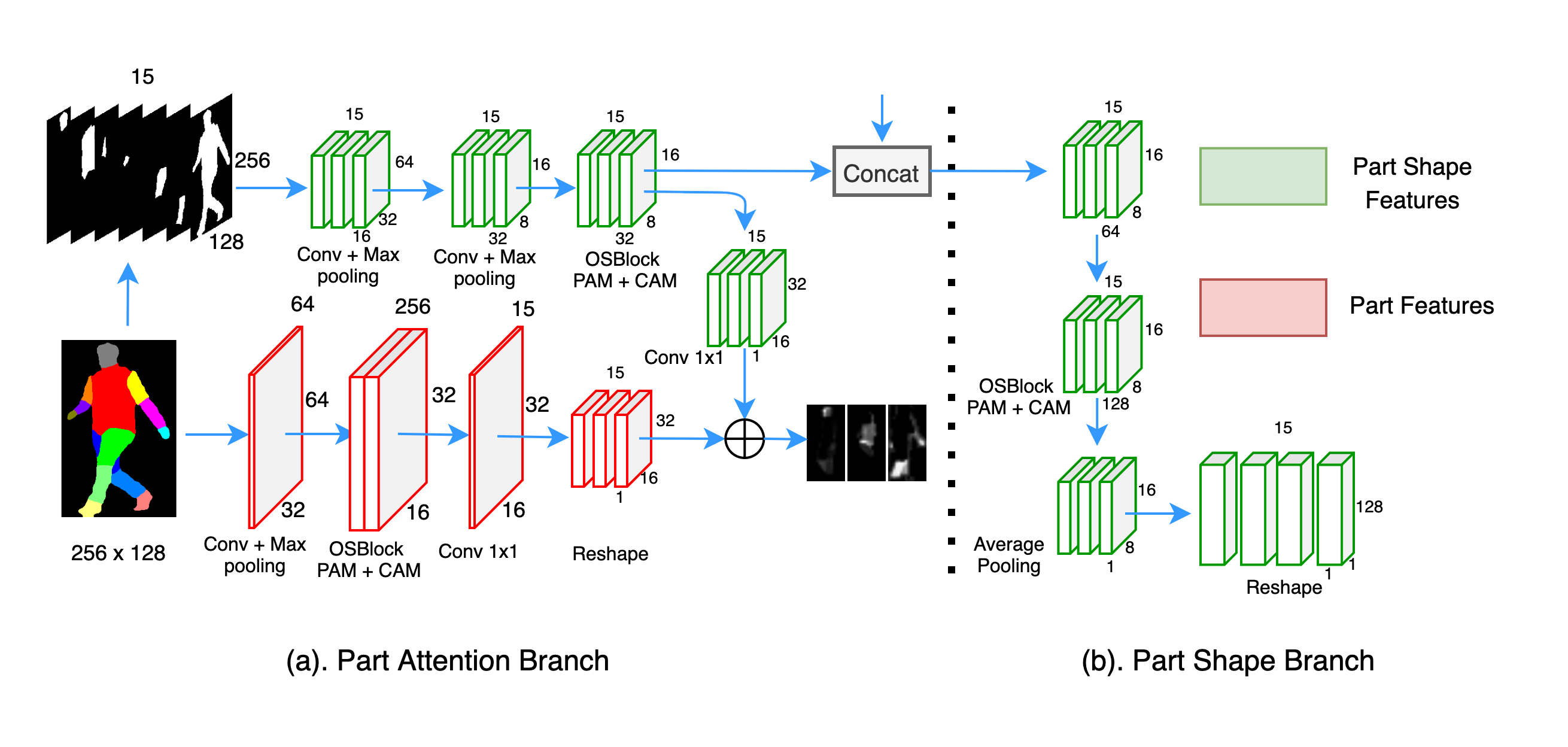}
\caption{\doublespacing The structure of part feature learning module. (a). In the part attention branch, part attention maps are jointly learned from features of human parsing maps and part parsing maps. (b). Features of part parsing maps of two images are concatenated for learning the part shape features in the part shape branch.}
\label{Part}
\end{figure*}

To the best of our knowledge, this is the first work to include a verification network for learning the specialized features. For a comprehensive understanding of the influence of network structure on this task, we investigate several widely used models\cite{PLR01,PCB01,ResNet01, Dense01} and establish our verification network using the same basic layer as the recently-proposed PLR-OSNet\cite{PLR01} to connect feature maps of different levels, i.e. the combination of an OSBlock\cite{OS01}, a Position Attention Module\cite{ABD01} and a Channel Attention Module\cite{CAM01}. With a better fitting to the human images, we propose our Local Clues Oriented Verification Network, which consists of three modules, global feature learning module, part feature learning module, and high order relation learning module. An overview of our LCVN is shown in \textcolor{blue}{Fig. \ref{Framework}}.

\textbf{Part Feature Learning} A verification network for similar images should be able to adaptively search and compare decisive clues among the input images. Nevertheless, it is meaningless to compare the features of a head with the features of feet across two images. For the convenience of hint locating, we apply patch-wise comparison to narrow the searching area. Specifically, we firstly parse each image into a human parsing map and 15 part parsing maps using the human parsing model\cite{CDCL01}. Human parsing map is a single image with each pixel a corresponding body part label while part parsing map is a binary image showing the area of the specific body part. To reach part-to-part matching, we learn an attention map for each body part from the human parsing map as well as the corresponding part parsing map, as shown in \textcolor{blue}{Fig. \ref{Framework}} and \textcolor{blue}{Fig. \ref{Part}(a)}. By incorporating the human parsing map, our part attention map is pose-aware and more robust to occlusions. We apply these attention maps to learn the part features of the input images. At last, following the early fusing principle, we fuse corresponding part features of two images as soon as they are established.

\textbf{Part Shape Feature Learning} Except for the body part features, the shape features are another crucial clues for discerning similar images. Though there are some recent works\cite{cc02,cc01,PRCC01} that learn body shape features for cloth changing Re-ID, these researches only work on the whole body silhouette, which is inadequate for similar images verification. In contrast, we involve part parsing maps in our model, which gives quick access to fine-grained part shape features comparison. To learn part shape features, we further extend our part feature branch by another part shape branch that compares the shape features of each body part in two input images (see \textcolor{blue}{Fig. \ref{Part}(b)}). Concatenating both the part features and the part shape features, our LCVN compares two similar images with well-designed features on a patch level.

\textbf{High Order Relation Learning} Although we have the features of different body parts, image verification is still challenging due to the similarity between the input images. Thus, it is necessary to exploit all possible features. To identify a person, the best choice is to rely on those unchangeable features. Just as one can change his/her color appearance by changing clothes, body/part shape can also be intentionally feigned by wearing loose clothes. We find that the ratio of limbs is another kind of unchangeable feature. Unfortunately, there are few reliable three-dimension human pose estimation models. As an alternative, we turn to the graph convolutional network (GCN) methods \cite{GCN01} to model the high-order information. To learn these inter-part features, we view each body part as a node and pass part features between these nodes using two adaptive directed graph convolutional(ADGC) layers. ADGC layer is proposed by Wang et al.\cite{HOReID01} for modeling the high-level information in occluded human images, which suits our case well. Restricted by pages, we do not show more details of the ADGC layer, please refer to the paper\cite{HOReID01}. We add a maximum pooling layer and a fully connected layer with batch normalization after the final ADGC layer so that the network can focus on  the most decisive clues.

\textbf{Global Feature Learning} When training on similar image pairs, the model is prune to become over-fitting. To overcome this problem, a global feature learning module is included in our LCVN for multi-target learning. We establish our global module with the same structure as PLR-OSNet\cite{PLR01} to verify two input images. Different from the part feature based main branch, our global module makes decisions only relying on two input images, which provides discriminating and reliable features for further part feature learning.

\subsection{Energy Function and Inference}
Though the target of the verification module is to know whether two input images are from the same person, during the inference stage, we care more about the relative similarities of the candidate images. Therefore, instead of using cross-entropy loss which targets at classification, triplet loss is adopted in this work because it sets loose restrictions on the output. For the same query image, we aim to learn relative similarity scores such that, positive candidate images are higher than positive candidate images, i.e.

\begin{center}
\begin{equation}
\mathcal{L}_{g} = \sum_{i = 1}^{N_{T}}\sum_{j = 1}^{P}\sum_{k = 1}^{P}[\mathrm{sim}_{G}(T_{i},T^{p}_{i_{j}^{F}}) - \mathrm{sim}_{G}(T_{i},T^{n}_{i_{k}^{F}}) - m]_{+} , [z]_{+} = max(z,0)
\end{equation}
\begin{equation}
\mathcal{L}_{p} = \sum_{i = 1}^{N_{T}}\sum_{j = 1}^{P}\sum_{k = 1}^{P}[\mathrm{sim}_{S}(T_{i},T^{p}_{i_{j}^{F}}) - \mathrm{sim}_{S}(T_{i},T^{n}_{i_{k}^{F}}) - m]_{+} , [z]_{+} = max(z,0)
\end{equation}
\begin{equation}
\mathcal{L} = \mathcal{L}_{g} + \mathcal{L}_{p}
\end{equation}
\end{center}

Here, $m$ is the margin between positive samples and negative samples. $\mathrm{sim}_{G}(\cdot,\cdot)$ and $\mathrm{sim}_{S}(\cdot,\cdot)$ are the similarity scores given by the global module and the high order relation learning module respectively. $\mathbf{T}_{i^{F}_{\cdot}}^{\cdot}$ is the similar image list of image $T_{i}$ given by the model $F$. $T^{p}_{i_{j}^{F}}$ is the $j$-$th$ positive image in the similar image list of $T_{i}$. $T^{n}_{i_{k}^{F}}$ is the $k$-$th$ negative image in the similar image list of $T_{i}$.

When inferring, we only take the score given by the high-order relationship learning module for similarity sorting. We adopt a novel ranking strategy that considers both the retrieval ranking result and the verification result. Each turn we only compare $L$ images according to the retrieval ranking. We left other images as substitutions for these $L$ images. Once the image with the highest similarity is taken, we fill the $L$ images set from substitution images according to their initial ranking. We keep doing like this until we take all $Q$ images. By selecting a proper $L$, our novel ranking strategy succeeds in controlling the risk of including too many dissimilar images.

\section{Experiments}

\subsection{Dataset}
Training and testing of the proposed architectures are conducted on publicly available VC-Clothes dataset\cite{VC01} and PRCC dataset\cite{PRCC01}.

VC-Clothes dataset is a synthetic dataset rendered by the video game engine Grand Theft Auto V with high Definition and realistic image effect. It is composed of 19,060 images of 512 identities. The dataset intentionally includes images from viewpoints where human faces can not be easily observed so it is imperative to identify persons by intrinsic clues like body shape and walking stance. Each identity appears in 4 scenes with 1 $\sim$ 3 suits of clothes. Suits are kept unchanged when captured in the same scene. We equally split the dataset by identities: 9449 images of 256 identities for training and 9611 images of the rest 256 identities for testing. A training set $\mathbf{T}$ with 202 identities and 7448 images is selected from the training segment. For the rest images in the training segment, 244 of them form the validation query set $\mathbf{VQ}$ and 1777 others form the validation gallery set $\mathbf{VG}$. Within the testing segment, 1020 images are used as queries $\mathbf{Q}$  and 8591 others form the gallery $\mathbf{G}$. We conduct experiments in the following two ways:
\begin{itemize}
    \item{\textbf{Full Setting}} All query images and gallery images are included for inference. However, since there are some images in the query set that do not have corresponding images with different clothes, we refine the number of images in the validation query set $\mathbf{VQ}$ to 162 and the number of images in the query set $\mathbf{Q}$ to 759.

    \item{\textbf{Partial Setting}} For the convenience of comparison, we also test our methods following the setting in \cite{VC01}. In this setting, only the query images from Cam-3 and the gallery images from Cam-4 are considered, which leads to a query set $\mathbf{Q}$ of 182 images and a gallery set $\mathbf{G}$ of 2258 images. Since this setting greatly limits the number of images in the query set and the gallery set, we still valid the model on the same dataset as in the full setting. 
\end{itemize}

Person Re-ID under moderate Clothing Change
(PRCC)\cite{PRCC01} dataset is a realistic dataset taken in an experimental setting with 3 cameras. PRCC dataset consists of 33698 images from 221 identities. Each person in Cameras A and B is wearing the same clothes, but the images are captured in different rooms. For Camera C, the person wears different clothes. The images in the PRCC dataset include not only clothing changes for the same person across different camera views but also other variations, such as changes in illumination, occlusion, pose, and viewpoint. PRCC datasets are split into three segments by identities: 18938 images of 124 identities as the training set $\mathbf{T}$, 3960 images of 26 identities for validation, and 10800 images of 71 identities for testing. All gallery images are included when inducing in the multi-shot setting. We treat images from camera A as the query set and images from camera C as the gallery set, which leads to a validation query set $\mathbf{VQ}$ of 1381 images, a validation gallery set $\mathbf{VG}$ of 1376 images, a query set $\mathbf{Q}$ of 3384 images and a gallery set $\mathbf{G}$ of 3543 images. 

\subsection{Implementation Details}
In the training phase, two networks are trained in sequence. Images are first fed into a candidate image retrieval network. SPRe-ID\cite{Sphere01}, PCB\cite{PCB01}, PLR-OSNet\cite{PLR01}, ABD-Net \cite{ABD01}, and FlipReID \cite{Flip01} are employed as the candidate image retrieval model $F$ for VC-Clothes dataset\cite{VC01} and PRCC dataset\cite{PRCC01}, with the default parameters and energy function. We form our new similar image sets with $P = 20$ candidate images for each query image.

In the second stage, our image verification model works on the new datasets for verifying similar image pairs. Images are padded and resized into 256 $\times$ 128 for the VC-Clothes dataset and 336 $\times$ 112 for the PRCC dataset. These images are further normalized and augmented with random horizontal flipping before entering the image verification network. Besides, we use CDCL\cite{CDCL01} for parsing images in both datasets. We pre-process the human parsing results in the same way as the RGB input images. To have our model work on the PRCC dataset, we insert a fully connected layer at the beginning of the first ADGC layer in the high-order relation learning module.

The batch size is set to 16 with 8 images per person for the VC-Clothes dataset and 12 with 6 images per person for the PRCC dataset. During the training stage, all three modules are jointly trained end-to-end for at most 80 epochs with the initialized learning rate 3.5e-4 and decaying to its 10\% at 30 and 60 epochs. 

For evaluation, we adopt the mean average precision (mAP) and rank-k accuracy(CMC). Unlike other works\cite{PRCC01,cc01}, which use single-shot matching by randomly choosing one image of each identity to form the gallery set $\mathbf{G}$, we work on the full gallery set and follow the standard Re-ID evaluation. When inferring the final results, we take two sets of $L$ and $Q$ for partial setting ($L=5,Q=10$) and full setting VC-Clothes ($L=10,Q=20$) respectively. We also test the effectiveness of our framework on PRCC dataset with two sets of $L$ and $Q$ (i.e. $L=10,Q=20$ and $L=20,Q=30$). The influence of these parameters are analyzed in Section 4.3 and 4.4. Generally speaking, they are decided by the number of similar images in the dataset. The computation efficiency of our algorithm is $\mathcal{O}(N_{Q})$ for inference. $N_{Q}$ is the number of images in the query set $\mathbf{Q}$. As for the time consumption, the average training time is about 50 hours for both VC-Clothes dataset and PRCC dataset with one NVIDIA GTX 1080 Ti GPU. It takes 0.33 seconds on average for processing each query image.

\subsection{Experimental Results}
We evaluate our proposed framework on two cloth changing datasets, i.e. VC-Clothes\cite{VC01} and PRCC\cite{PRCC01}. Since the main target of this work is to research the effectiveness of specialized features in processing similar candidate images, we test our framework on several representative retrieval models. \par
\begin{table}[h]
    \caption{\doublespacing \textbf{Quantitative results on VC-Clothes dataset.} Four large rows show the results of hand-crafted features approaches, global features based method, part features base models, and re-rank methods, respectively. “R-k” denotes rank-k accuracy (\%). “mAP” denotes mean average precision (\%). “-” denotes not reported.}
    \centering
    {\small \begin{tabular}{l|c|c|c|c||c|c|c|c}
    \hline
    \hline
      \multirow{2}{*}{Methods}  & \multicolumn{4}{c||}{Partial Setting $L = 5, Q = 10$} & \multicolumn{4}{c}{Full Setting $L = 10, Q = 20$ } \\ \cline{2-9}
         & R-1 & R-5 & R-10 & mAP &  R-1 & R-5 & R-10 & mAP \\
    \hline 
      LOMO+XQDA\cite{LOMO01}  & 34.5 & 44.2 & 49.7 & 30.9 & - & - & - & - \\
      GOG\cite{HOG01}+XQDA\cite{LOMO01}  & 35.7 & 48.3 & 54.2 & 31.3  & - & - & - & - \\
    \hline 
    \hline
      MDLA\cite{MDLA01}  & 59.2 & 67.3 & 73.5 & 60.8  & - & - & - & -\\
      Dense-Net\cite{Dense01} & - & - & - & - & 73.1 & 79.3 & 82.7 & - \\
      NAS\cite{NAS01} & - & - & - & -  & 77.4 & 84.1 & 87.1 & - \\
      MHO\cite{MHO01} & - & - & - & - & 57.1 & 61.6	& 64.2 & 52.5\\
      DG-Net\cite{DGNet01} & 75.8 & 80.8 & 83.5 & 70.3 & 80.5 & 86.2	& 90.1 & 73.0\\
      FlipReID\cite{Flip01}(\textit{baseline1})  & 82.4 & 90.7 & 94.0 & 85.5 & 86.2 & 90.7 & 93.1 & 81.2 \\
      ABD-Net\cite{ABD01}(\textit{baseline2})  & 83.0 & 88.5 & 90.7 & 84.0 & 87.0 & 91.2 & 94.2 & 81.1\\
    \hline
    \hline
      SPReID\cite{Sphere01}(\textit{baseline3}) & 40.7 & 54.4 & 62.1 & 33.8 & 42.7 & 54.8 & 62.7 & 29.4 \\
      FSAM\cite{Part01}  & 78.6 & - & - & 78.9  & - & - & - & -\\
      HOReID\cite{HOReID01}   & - & - & - & - & 80.4 & 88.0 & 91.0 & 73.6\\
      PCB\cite{PCB01}(\textit{baseline4})  & 64.3 & 75.8 & 81.3 & 63.2 & 66.4 & 77.9 & 83.7 & 56.9 \\
      PLR-OSNet\cite{PLR01}(\textit{baseline5})  & 82.4 & 89.6 & 92.3 & 81.8  & 84.1 & 89.5 & 92.5 & 76.0\\
    \hline
      FlipReID(\textit{baseline1})  & 82.4 & 90.7 & 94.0 & 85.5 & 86.2 & 90.7 & 93.1 & 81.2 \\
       + Re-Rank\cite{Rank01}  & 82.4 & 90.7 & 94.0 & 85.5 & 86.2 & 90.7 & 93.1 & 81.2 \\
       + LCVN  & \textbf{85.7}  & \textbf{91.8} & 94.0 & \textbf{85.7} &  \textbf{89.3} & \textbf{92.2} & \textbf{93.8} & \textbf{81.3} \\ 
       + Re-Rank + LCVN  & \textbf{85.7}  & \textbf{91.8} & 94.0 & \textbf{85.7} &  \textbf{89.3} & \textbf{92.2} & \textbf{93.8} & \textbf{81.3} \\ 
      \hline
      ABD-Net(\textit{baseline2})  & 83.0 & 88.5 & 90.7 & 84.0 & 87.0 & 91.2 & 94.2 & 81.1 \\
       + Re-Rank  & 85.7 & 89.6 & \textbf{91.7} & \textbf{86.7} & 87.3 & 91.6 & 93.7 & 84.0 \\
       + LCVN   & 84.6 & 90.1 & 90.7 & 84.0  &  90.7 &  \textbf{94.2} & 95.4 & 81.9 \\ 
       + Re-Rank + LCVN & \textbf{87.9}  & \textbf{91.2} & \textbf{91.7} & 86.6 &  \textbf{91.7} & \textbf{94.2} & \textbf{95.5} & \textbf{84.3} \\ 
      \hline
      SPReID(\textit{baseline3})  & 40.7 & 54.4 & 62.1 & 33.8 & 42.7 & 54.8 & 62.7 & 29.4 \\
       + Re-Rank  & 47.8 & 53.8 & 62.1 & 48.9 & 52.3 & 56.9 & 61.9 & 44.3 \\
       + LCVN  & 46.7 & \textbf{58.2} & 62.1 & 35.9  &  56.5 & 66.3 & \textbf{71.3} & 33.0 \\ 
       + Re-Rank + LCVN & \textbf{51.6}  & \textbf{58.2} & 62.1 & \textbf{49.3} &  \textbf{58.7} & \textbf{67.1} & 70.7 & \textbf{45.5} \\ 
      \hline
      PCB(\textit{baseline4})  & 64.3 & 75.8 & 81.3 & 63.2 & 66.4 & 77.9 & 83.7 & 56.9 \\
       + Re-Rank  & 65.4 & 74.7 & 81.3 & \textbf{66.7} & 70.4 & 76.7 & 83.1 & 63.8 \\
       + LCVN  & 67.6 & \textbf{78.6} & 81.3 & 63.7  &  75.0 &  \textbf{83.9} & \textbf{87.7} & 58.6 \\ 
       + Re-Rank + LCVN & \textbf{69.2}  & 77.5 & 81.3 & \textbf{66.7} &  \textbf{77.5} & 83.8 & \textbf{87.7} & \textbf{64.4} \\
      \hline
      PLR-OSNet(\textit{baseline5})  & 82.4 & 89.6 & 92.3 & 81.8 & 84.1 & 89.5 & 92.5 & 76.0 \\
       + Re-Rank  & 82.4 & 90.1 & \textbf{94.0} & 84.3 & 85.0 & 89.3 & 92.8 & \textbf{79.2} \\
       + LCVN  & 84.6 & 91.8 & 92.3 & 81.9  &  \textbf{87.7} &  91.8 & 93.5 & 76.6 \\ 
       + Re-Rank + LCVN & \textbf{85.2}  & \textbf{92.3} & \textbf{94.0} & \textbf{84.6} &  \textbf{87.7} & \textbf{92.9} & \textbf{93.7} & 79.1 \\ 
      \hline
    \end{tabular}
    }
    \label{VC1}
\end{table}

\textbf{Result on VC-Clothes} We compared our method with several representative state-of-the-art methods including hand-crafted feature based methods\cite{LOMO01,HOG01}, global features based Re-ID methods\cite{ABD01,Dense01,Flip01,MDLA01,NAS01,MHO01,DGNet01}, part features based Re-ID\cite{PLR01,PCB01,Part01,HOReID01}, and re-rank methods\cite{Rank01}. The experiment results on the VC-Clothes dataset are shown in \textcolor{blue}{Table \ref{VC1}}. As one can see, there is a big performance gap between hand-crafted feature based methods and deep-learning based methods. While within deep-learning based methods, vanilla global feature based methods\cite{ABD01,Flip01} and part feature based deep learning methods\cite{PLR01} achieve similar performance.  FSAM\cite{Part01} is a method specially designed for the cloth changing Re-ID problem. The biggest difference between FSAM and our work is that we take local clues as specialized features while they treat them as general features. One can find that FSAM performs even worse than some vanilla Re-ID models. These experiment results imply the effectiveness of the existing Re-ID methods on general feature extraction as well as the potential of local clues as specialized features. 

\afterpage{%
   \clearpage 
    \vspace{-3cm}
\begin{table}[t]
    \centering
    \caption{ \textbf{Quantitative results on the PRCC dataset.\cite{PRCC01}} Four large rows show the results of hand-crafted approaches, global features based methods, part features based methods, and re-rank methods respectively.  $\ast$ means that the result is achieved on a smaller gallery set.}
    {\small
    \begin{tabular}{l|c|c|c|c}
    \hline
    \hline
      Methods  & R-1 & R-10 & R-20 & mAP\\
    \hline 
      LOMO+XQDA\cite{LOMO01} & $14.5^{\ast}$ & - & - & -\\
      HOG\cite{HOG01}+XQDA\cite{LOMO01} & $22.1^{
      \ast}$ & - & - & -\\
    \hline 
    \hline
      Alexnet\cite{Alex01} & $16.3^{\ast}$ & - & - & -\\
      VGG16\cite{VGG01} & $18.2^{\ast}$ & - & - & - \\
      Res-Net50\cite{ResNet01} & $19.4^{\ast}$ & - & - & -\\
      HA-CNN\cite{HACNN01}& $21.8^{\ast}$ & -	& - & -\\
      MHO\cite{MHO01} & 20.9 & 35.6 & 41.7 &  19.5\\
      CASE-Net\cite{cc02} & $39.5^{\ast}$ & - & - & -\\
      DG-Net\cite{DGNet01} & 47.9 & 54.9 & 58.9 & 46.9 \\
      ABD-Net\cite{ABD01}(\textit{baseline1}) & 49.3 & 58.7 & 62.4 & 44.8\\
      RCSANet\cite{Overall01} & $50.2^{\ast}$ & - & - & $48.9^{\ast}$ \\
      LAST\cite{LAST01} & $57.5^{\ast}$ & - & - & $54.7^{\ast}$ \\
      FlipReID\cite{Flip01}(\textit{baseline2}) & 62.2 & 69.8 & 72.5 & 62.3 \\
    \hline
    \hline
      SPReID\cite{Sphere01}(\textit{baseline3}) & 21.1 & 31.6 & 35.7 & 19.8 \\
      STN\cite{STN01} & $27.5^{\ast}$ & - & - & -\\
      SPT + ASE\cite{PRCC01} & $34.4^{\ast}$ & - & - & -\\
      PCB\cite{PCB01}(\textit{baseline4}) & 39.6 & 47.8 & 49.3 & 37.8\\
      PLR-OSNet\cite{PLR01}(\textit{baseline5}) & 49.6 & 59.4 & 64.1 & 48.5\\
      HOReID\cite{HOReID01} & 43.8 & 54.1 &56.5 & 43.8\\ 
      FSAM\cite{Part01} & $54.5^{\ast}$ & - & - & - \\
    \hline
    \hline
      ABD-Net(\textit{baseline1}) & 49.3 & 58.7 & 62.4 & 44.8\\ 
       + Re-rank(\cite{Rank01}) $L = 10, Q = 20$ & 49.3 & 58.7 & 62.4 & 44.8\\ 
       + LCVN $L = 10, Q = 20$ & \textbf{51.7} & 60.8 & 62.4 & 44.8\\ 
      + Re-rank + LCVN $L = 10, Q = 20$ & \textbf{51.7} & 60.8 & 62.4 & 44.8\\ 
       + LCVN $L = 20, Q = 30$ & 50.6 & \textbf{63.2} & \textbf{64.1} & \textbf{45.0}\\ 
       + Re-rank + LCVN $L = 20, Q = 30$ & 50.6 & \textbf{63.2} & \textbf{64.1} & \textbf{45.0}\\ 
    \hline
      FlipReID(\textit{baseline2}) & 62.2 & 69.8 & 72.5 & 62.3\\ 
       + Re-rank & 62.2 & 69.8 & 72.5 & 62.3\\ 
       + LCVN $L = 10, Q = 20$  & \textbf{64.4} & 70.8 & 72.5 & \textbf{62.6}\\ 
      + Re-rank + LCVN $L = 10, Q = 20$ & \textbf{64.4} & 70.8 & 72.5 & \textbf{62.6}\\ 
       + LCVN $L = 20, Q = 30$ & 64.2 & \textbf{71.6} & \textbf{73.5} & \textbf{62.6}\\ 
      + Re-rank + LCVN $L = 20, Q = 30$ & 64.2 & \textbf{71.6} & \textbf{73.5} & \textbf{62.6} \\ 
    \hline
      SPReID(\textit{baseline3}) & 21.1 & 31.6 & 35.7 & 19.8\\ 
      + Re-rank & 21.1 & 31.6 & 35.7 & 19.9\\ 
       + LCVN $L = 10, Q = 20$ & 26.2 & 33.6 & 35.7 & 20.3\\ 
       + Re-rank + LCVN $L = 10, Q = 20$ & 26.3 & 33.6 & 35.7 & 20.3\\ 
       + LCVN $L = 20, Q = 30$ & \textbf{27.8} & 34.9 & \textbf{37.0} & 20.6\\ 
       + Re-rank + LCVN $L = 20, Q = 30$ & \textbf{27.8} & \textbf{35.0} & \textbf{37.0} & \textbf{20.7} \\ 
    \hline
      PCB(\textit{baseline4}) & 39.6 & 47.8 & 50.8 & 42.0\\ 
      + Re-rank(\cite{Rank01}) & 40.0 & 48.3 & 50.8 & 40.4\\ 
      + LCVN $L = 10, Q = 20$ & 43.6 & 49.1 & 52.1 & 42.7\\ 
      + Re-rank + LCVN $L = 10, Q = 20$ & 43.8 & 50.1 & 52.1 & 40.8\\ 
      + LCVN $L = 20, Q = 30$ & 44.9 & 50.2 & 52.5 & \textbf{42.8}\\ 
       + Re-rank + LCVN $L = 20, Q = 30$ & \textbf{45.0} & \textbf{51.7} & \textbf{53.9} & 40.9\\ 
    \hline
      PLR-OSNet(\textit{baseline5}) & 49.6 & 59.4 & 64.1 & 48.5\\ 
       + Re-rank & 49.5 & 60.1 &  64.5 & 49.6\\ 
       + LCVN $L = 10, Q = 20$ & 52.2 & 61.6 & 64.1 & 48.6\\ 
       + Re-rank + LCVN $L = 10, Q = 20$ & 53.6 & 62.1 & 64.5 & \textbf{49.8}\\ 
       + LCVN $L = 20, Q = 30$ & 53.3 & 63.0 & 65.3 & 48.6\\ 
       + Re-rank + LCVN $L = 20, Q = 30$ & \textbf{54.3} & \textbf{63.2} & \textbf{65.5} & \textbf{49.8} \\
    \hline
    \end{tabular}
    }
    \label{PRCC1}
    \vspace{-3cm}
\end{table}
}
\textcolor{white}{ }\par
\textcolor{white}{ }\par
\textcolor{white}{ }\par
\textcolor{white}{ }\par
\textcolor{white}{ }\par
We also compare our method with other re-rank method\cite{Rank01}. Strictly speaking, our verification network can be viewed as a kind of re-rank strategy. Compared with \cite{Rank01}, our method achieves more improvements on the CMC metrics but fewer improvements on the mAP metrics. This is because our method is designed for handling similar images in cloth changing Re-ID datasets and can consequently affect only a few similar images. On the contrary, re-ranking all images according to the sample-wise feature distances can help find more other potential images, but it can hardly affect those similar images because they usually have similar general features. From this viewpoint, two re-rank strategies are complementary to each other. The experiment results also reflect this point. Combining two re-rank methods, impressive improvement can be achieved on all evaluation metrics.

\textbf{Result on PRCC}
We also execute a comprehensive experiment to evaluate the performance of our proposed model on the realistic cloth changing Re-ID dataset, PRCC. Due to the difference in the problem definition, i.e. a smaller gallery set is selected for evaluation in the original paper of PRCC\cite{PRCC01}, we only keep the Rank-1 result for the model tested in \cite{PRCC01}. We further include some other Re-ID methods to evaluate the effectiveness of our framework on the realistic cloth changing dataset. The results are shown in \textcolor{blue}{Table \ref{PRCC1}}.  

Our proposed method achieves the best Rank-1 accuracy up to 64.4\% among other compared methods, including hand-crafted features methods\cite{LOMO01,HOG01}, global features based methods\cite{cc02,LAST01,ABD01,Overall01, ResNet01,Flip01, MHO01,Alex01,VGG01,HACNN01,DGNet01}, part features based models\cite{PLR01,PCB01,Part01,STN01,PRCC01,HOReID01,Sphere01} and re-rank methods\cite{Rank01}. The performance on the PRCC dataset indicates its difficulty. Some of these Re-ID models\cite{PLR01,ABD01,MHO01} achieve more than 95\% Rank-1 accuracy on the cloth-consistent dataset\cite{mar01}, but only achieve less than 50\% Rank-1 accuracy in the cross clothes matching test. \par

The performance of the tested methods shares a similar trend with the result on VC-Clothes. Both part feature based methods and cloth changing Re-ID models achieve competitive results on the PRCC dataset. On the other hand, LAST\cite{LAST01} and FlipReID\cite{Flip01} learn general features without the help of part feature learning and outperform other methods by a large margin. This phenomenon can also be observed when testing on the VC-Clothes dataset, suggesting that the part features are not necessary to be general features in the cloth changing Re-ID case.  \par

Re-rank method\cite{Rank01} is also compared in this experiment. However, in some cases, no significant changes in evaluation metrics can be observed after applying re-ranking. We check the calculation of the re-ranked feature distance matrix and find that almost all neighbor images are reciprocal in these methods. This means the Jaccard distance is 1 for almost every pair of images. Adding the same number on the feature distance matrix makes no change to the final Re-ID result. The occurrence of this phenomenon is affected by many factors, including the whole data distribution, the similarity between images in the dataset, the selection of retrieval model, etc. From this viewpoint, our retrieval-verification framework is a more robust and stable re-rank strategy.

\begin{table}[h]

    \caption{\doublespacing Analysis of global feature learning module $\mathcal{G}$, part feature learning module $\mathcal{P}$ and high-order relation learning module $\mathcal{H}$ on VC-Clothes dataset and PRCC dataset.}
    \centering
    \begin{tabular}{c|c|c|c|c|c||c|c|c}
    \hline
    \hline
    \multicolumn{3}{c|}{} & \multicolumn{3}{c||}{VC-Clothes} & \multicolumn{3}{c}{PRCC} \\ \hline
      $\mathcal{G}$ & $\mathcal{P}$ & $\mathcal{H}$ & R-1 & R-5 & R-10 & R-1 & R-5 & R-10 \\
    \hline 
      $\checkmark$ &  & & 90.9 & 93.5 & 94.8 & 62.5 & 67.3 & 69.8 \\
    \hline
      $\checkmark$ & $\checkmark$ &  & 91.2 & 93.8 & 95.3  & 63.4 & 67.9 & 70.3 \\
    \hline
      $\checkmark$ & $\checkmark$ & $\checkmark$ & \textbf{91.7} & \textbf{94.2} & \textbf{95.5} & \textbf{64.4} & \textbf{68.9} & \textbf{70.8}\\
    \hline
    \hline
    \end{tabular}
    \label{Ablation_Module1}
\end{table}

\subsection{Ablation Study}

To analyze the influence of different components in our framework, we include the results of comprehensive ablative experiments in this section. All experiments in this section are conducted on both the full setting VC-Clothes dataset and the PRCC dataset. The experiments on dataset VC-Clothes and PRCC are done using retrieval model ABD-Net\cite{ABD01} and FlipReID\cite{Flip01} respectively, for their competitive performances on the corresponding dataset. We set parameters $L=10$ and $Q=20$.

\textbf{Ablation Study on Model Design} In our LCVN, three modules are incorporated, i.e. the global feature learning module $\mathcal{G}$, the part feature learning module $\mathcal{P}$, and the high order relation learning module $\mathcal{H}$. The results are shown in \textcolor{blue}{Table \ref{Ablation_Module1}}. Comprehensive ablation experiments are conducted to analyze the effectiveness of each module. Firstly, we remove all other modules and only keep a global module as our basic model. We gradually add part feature learning module and high-order relation learning module to our model. 0.9\% and 1.9\% performance boosts are observed on the VC-Clothes and PRCC datasets respectively. The experiment results show how auxiliary features help the verification model to discern similar images. Compared with other part feature based methods, learning specialized auxiliary features can better activate their potential on verifying unwanted images.

\begin{table}[h!]
    \caption{\doublespacing Analysis of different loss functions on VC-Clothes dataset and PRCC dataset.}
    \centering
    \begin{tabular}{c|c|c|c||c|c|c}
    \hline
    \hline
    \multirow{2}{*}{} & \multicolumn{3}{c||}{VC-Clothes} & \multicolumn{3}{c}{PRCC} \\ \cline{2-7}
     & R-1 & R-5 & R-10 & R-1 & R-5 & R-10\\
    \hline 
    Cross Entropy & 91.2 & 93.7 & 95.1 & 63.6 & 68.0 & 70.1\\
    \hline 
    Cross Entropy + Circle Loss & 90.6 & 93.6 & 94.8 & 63.2 & 67.8 & 69.9 \\
    \hline
    Cross Entropy + Triplet Loss & 90.9 & 93.8 & 95.0 & 63.0 & 67.9 & 70.0 \\
    \hline
    Cross Entropy + Cosine Face Loss & 90.4 & 93.5 & 94.7 & 62.7 & 67.4 & 70.0 \\
    \hline
    Cross Entropy + Sphere Face Loss\cite{Sphere01}  & 90.1 & 93.3 & 95.3 & 62.6 & 67.2 & 70.2 \\
    \hline
    Triplet Loss & \textbf{91.7} & \textbf{94.2} & \textbf{95.5} & \textbf{64.4} &  \textbf{68.9} &\textbf{70.8}\\
    \hline
    \hline
    \end{tabular}
    \label{Loss}
\end{table}

\textbf{Ablation Study on Loss Function}
To deal with the image verification problem, one can treat it as a classification problem or a regression problem, depending on the selection of loss function. The main difference is whether we should set the same learning target for all same identity image pairs. If we take it as a classification problem, the output of our verification network is a confidence score representing whether two images are from the same person. In this case, the outputs of all same identity image pairs are supposed to be same. If we understand it as a regression problem, the labels will only work as an indicator of relative similarity. The responsibility of the verification network will also be changed to judge the relative similarity of the input images. This will allow the same identity image pairs of different query images have different learning targets. Therefore, it is a relatively loose setting.

In this experiment, we test how different loss functions affect the performance of our model. The results are shown in \textcolor{blue}{Table \ref{Loss}}. We treat this problem as a regression problem when testing with triplet loss and treat it as a classification problem with other loss functions. This is decided by the property of the loss functions because only triplet loss can work without setting a fixed similarity score learning target for the input image pairs. The experiment results suggest that it is more suitable to treat the verification problem as a regression problem. Besides, a restrict setting of margin between all positive samples and negative samples does not lead to better performances in this problem. This may be because  similarities of positive image pairs are be different for different identities. Higher flexibility of the output spaces can better fit the real distributions of image-wise similarity.

\begin{figure}[h!]
\centering
\vspace{0.5cm}
\includegraphics[width=\linewidth]{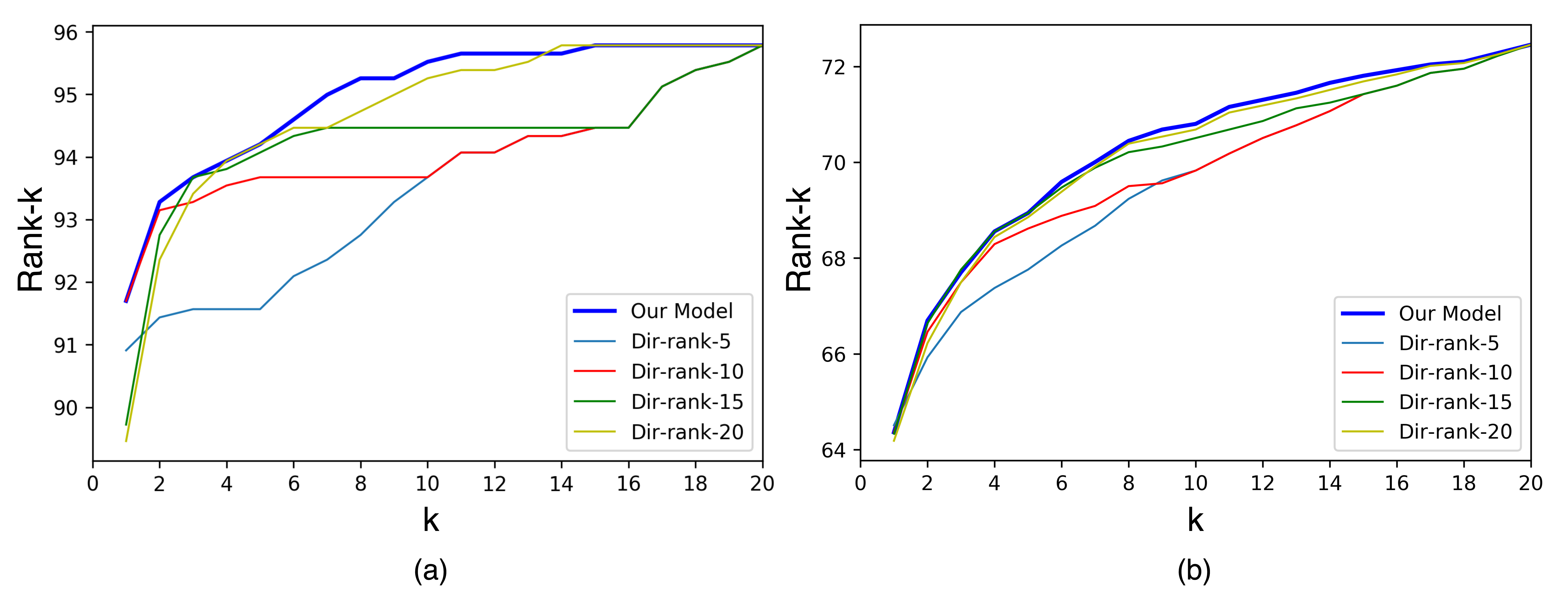}
\caption{\doublespacing Comparison on ranking strategy. Ranking results using our ranking strategy and direct re-rank strategy are compared on full setting VC-Clothes(a) and PRCC dataset(b).}
\label{Strategy}
\end{figure}

\textbf{Ablation Study on Ranking Strategy}  As explained in chapter \texttt{4.3}, we adopt a novel ranking strategy that considers both the retrieval ranking results and the verification ranking results when inferring. In this experiment, we evaluate our proposed ranking strategy by comparing the CMC curves given by our ranking strategy and traditional methods which directly sort the scores of top-5, 10, 15, and 20 images, as shown in \textcolor{blue}{Fig. \ref{Strategy}}. In this experiment, $L$ is set as default, i.e. $L = 10$. As an intuitive comparison, our ranking strategy outperforms other ranking strategies by a large margin in almost all Rank-$k$ metrics. We can also learn how the balance between considering more images and reducing the risk affects the verification result from this figure. Our ranking strategy takes a good balance on these points and achieves the biggest Area Under Curve(AUC) among these methods.

\begin{figure}[h]
\centering
\includegraphics[width=0.7\linewidth]{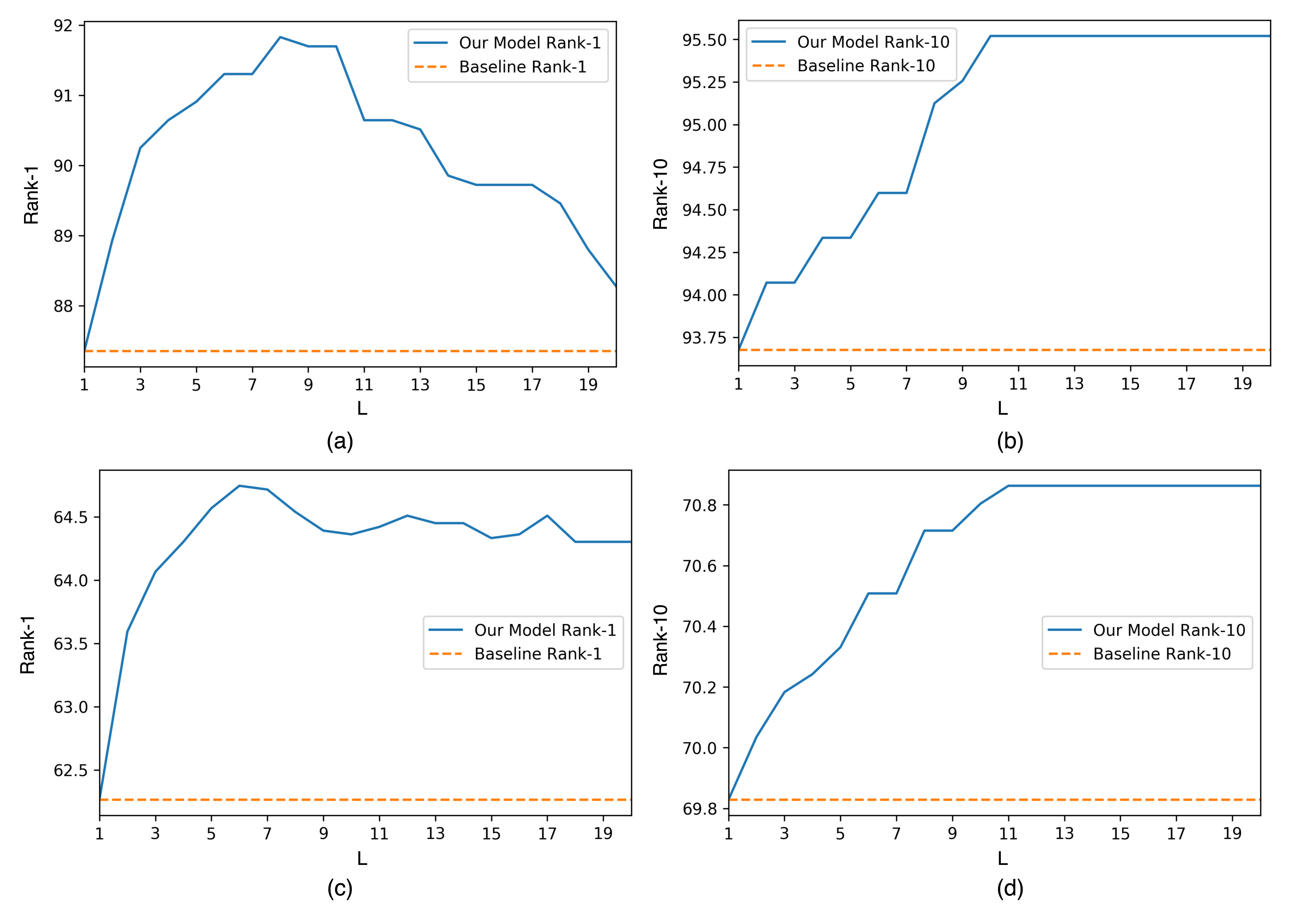}
\caption{\doublespacing (a,b). The influence of $L$ on Rank-1 and Rank-10 metrics on VC-Clothes\cite{VC01} dataset. (c,d). The influence of $L$ on Rank-1 and Rank-10 metrics on PRCC\cite{PRCC01} dataset.}
\label{selL}
\end{figure}

We also conduct experiments to analyze the influence of the selection of $L$, as shown in \textcolor{blue}{Fig. \ref{selL}}. We evaluate the performance of our model under different settings of $L$ on both datasets. \textcolor{blue}{Fig. \ref{selL}(a,c)} shows the influence of $L$ on the Rank-1 metric results when testing on the VC-Clothes dataset\cite{VC01} and PRCC dataset\cite{PRCC01}. As one can easily see, an increase in $L$ will include more images for comparison, so that the wanted image will have a higher possibility to be included and found. On the other hand, it is also accompanied by many unrelated images. Therefore, when working on the Rank-1 metrics, the ability to find the most potential image is evaluated. Instead, the Rank-10 metrics can better reflect the ability to eliminate dissimilar images. The verification network must find $L$ most dissimilar images among $L+10$ candidate images to get a satisfying Rank-10 result. However, since we only compare top $Q = 20$ gallery images for each query image, the increase in $L$ does not affect the Rank-10 results when $L$ is over 10.(see \textcolor{blue}{Fig. \ref{selL}(b,d)}). Experiment results on both datasets suggest that setting $L$ around 10 can reach the best performance. Even though, the "optimal" $L$ is greatly affected by the property and the scale of the dataset and is still underresearched.

\subsection{Further Analysis}

\begin{figure}[h]
\centering
\includegraphics[width=.8\linewidth]{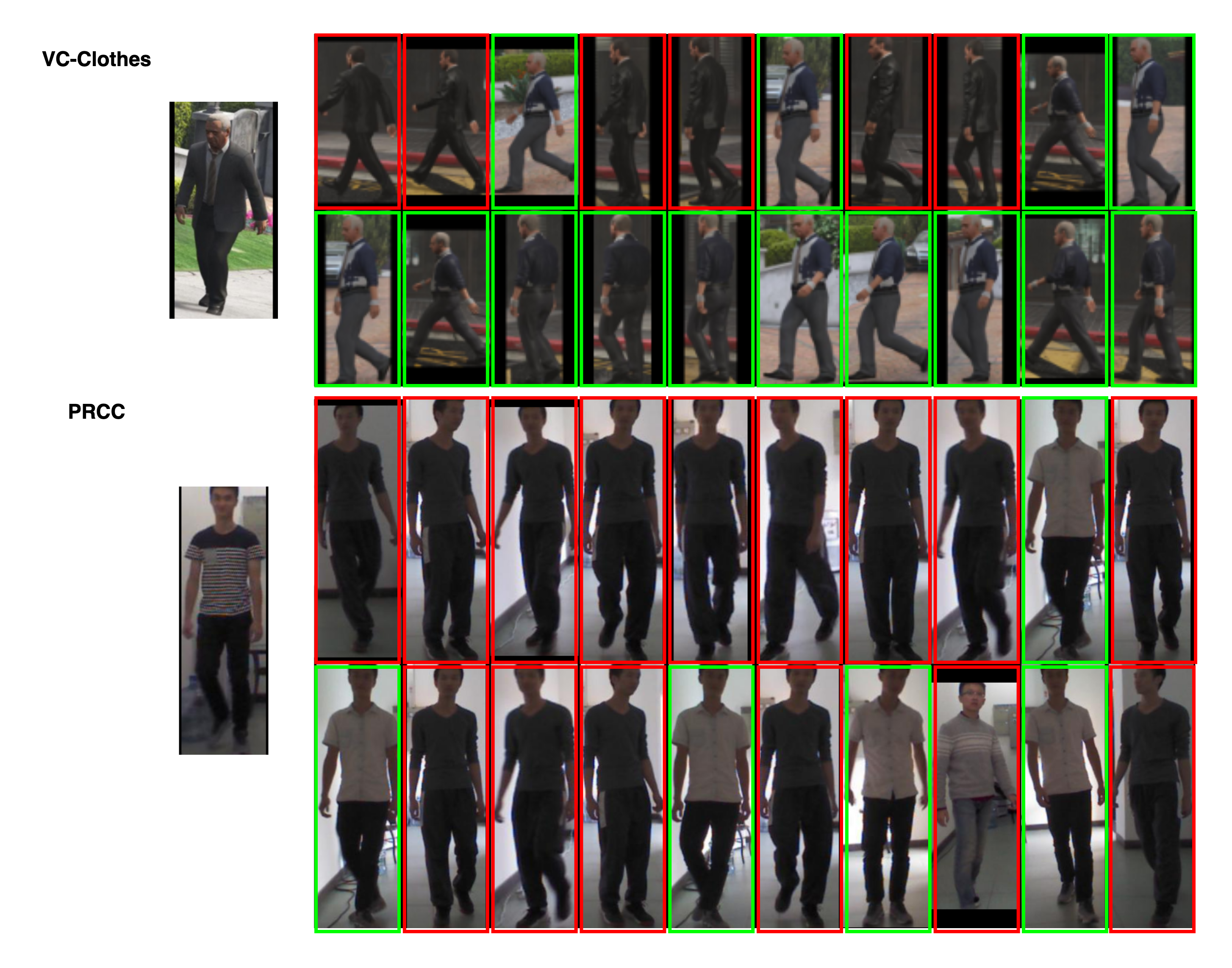}
\caption{\doublespacing The visualization of new ranking results by our proposed verification model on VC-Clothes and PRCC datasets. The first column consists of query images. For each query image, 10 most similar gallery images before and after the verification module are illustrated. Person surrounded by green box denotes the same person as the query image. Person surrounded by red box denotes the different person as the query image.}
\label{Qual_VC}
\end{figure}

\textbf{Visualized Verification Results} To give an intuitive evaluation of our framework, we visualize the candidate images before and after verification. Qualitative results of our model on the VC-Clothes dataset and PRCC dataset are shown in \textcolor{blue}{Fig. \ref{Qual_VC}}. Images in the first column are the query images from the query set $\mathbf{Q}$. For each query image, two rows with 10 images per row are included, representing the 10 most similar images before and after the image verification module. Just as suggested in the quantitative tests, our verification model is able to give a better ranking result through pair-wise image comparison. In most cases, the new ranking given by our verification network outperforms the result of the vanilla ReID model. Moreover, the qualitative testing results on both datasets suggest the importance of the lighting condition and the image resolution. We find that in most failed cases, query images are taken in bad lighting conditions or with low image quality. These factors will greatly affect the result of human parsing as well as the ID-related stable feature extraction. 

\begin{figure}[h!]
\centering
\includegraphics[width=.85\linewidth]{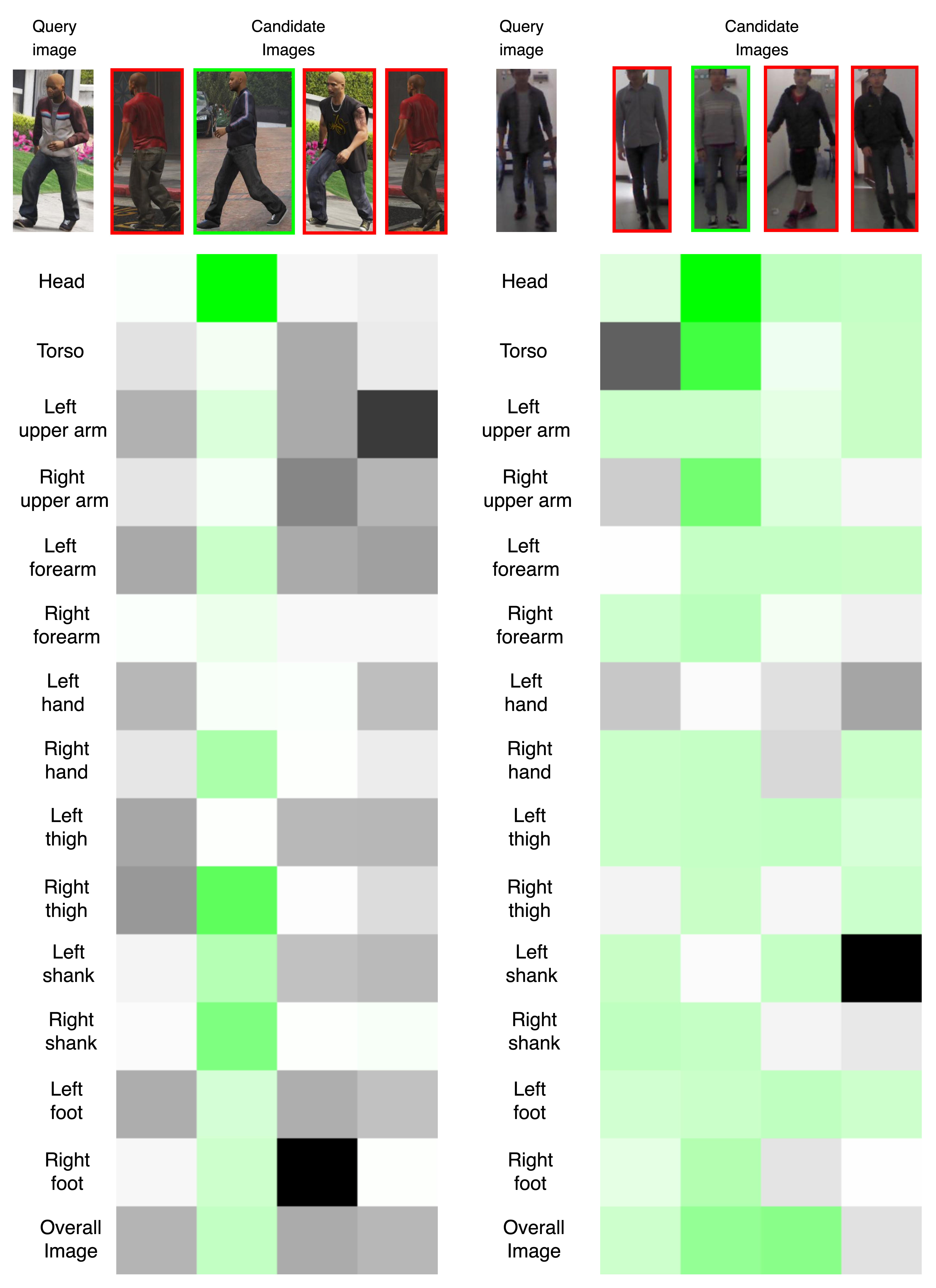}
\caption{\doublespacing The visualization of the contribution of different part features to the final similarity score. Green color indicates positive contributions and gray color indicates negative contributions. The left part is the results from VC-Clothes dataset. The left part is the results from PRCC dataset.}
\label{Part_contribution}
\end{figure} 

\textbf{How verification network views similar images} To understand how our verification network makes decisions, we extract the high-order part features produced by module $\mathcal{H}$ and calculate their contributions to the final score. The results are visualized in \textcolor{blue}{Fig. \ref{Part_contribution}}. The left part of the figure is the results from VC-Clothes dataset and the right part is the results from PRCC dataset. The color of each block represents the contribution of each part features to the final score. Green color indicates positive contributions and gray color indicates negative contributions. Since our verification network only works for inferring the relative similarity between images, the visualization result also just indicates the relative contribution of each body part instead of the absolute similarity. From the figure, we can learn that our verification network is able to flexibly arrange its attention according to the input image pair. For example, when comparing the query image with the first candidate image in PRCC, our LCVN pays more attention to the upper part of the image because the height of the shoulder may be decisive in this pair of comparisons. When comparing the query image with the last candidate image in PRCC, the bottom part of the image becomes more important because the leg strength in the two images seems different. Since we inserted a maximum pooling layer in the module $\mathcal{H}$, only the most decisive clues are considered. Even though, there are still many candidate images that we can not clearly tell how they are different from the query image(e.g. the third candidate image in PRCC). For these images, the best method is to find the most similar image among them. Our LCVN also handles this job well. The result of the second pair of comparisons in PRCC outperforms the third pair of comparisons in almost all body parts, suggesting the effectiveness of our methods.

\section{Conclusion}
In this paper, we focus on a challenging yet significant person re-identification task that has been long ignored in the past. We point out that the flexibility on color appearance introduces more similar images and details play a more crucial role in this task than before. Therefore, we treat the cloth changing person re-identification problem as the combination of an image retrieval task and a pair-wise image verification network. We present a novel Local Clues oriented Verification Network(LCVN) that works on similar samples in the dataset. Besides, in this work, we introduce a novel ranking strategy that takes a good balance between the initial retrieval results and new image similarity scores. By conducting extensive experiments, we demonstrate that the effectiveness of the two-step retrieval-verification framework as well as the proposed image verification model. In general, this paper tries to tackle the cloth changing person re-identification problem from the viewpoint of pair-wise image verification, which is rare in the recent Re-ID community. We believe this work will propel the development of the research of cloth changing Re-ID. 




%



\ifCLASSOPTIONcaptionsoff
  \newpage
\fi



%
\bibliographystyle{IEEEtran}
\bibliography{references}

%




\end{document}